\documentclass[lnbip]{svmultln}

\usepackage{makeidx}  
\usepackage{graphicx}

\usepackage{amsfonts}
\usepackage{xcolor}

\begin{document}
\mainmatter              
%
\title{Using Reinforcement Learning to Optimize Responses in Care Processes: A Case Study on Aggression Incidents}
\titlerunning{Using Reinforcement Learning to Optimize Processes}  
%
\author{Bart J. Verhoef\inst{1} \and
Xixi Lu\inst{1}
}%
\authorrunning{Verhoef et al.}   
%
%
\institute{Utrecht University, Utrecht, the Netherlands,\\
\email{b.j.verhoef@students.uu.nl, x.lu@uu.nl}
}

\maketitle              

\begin{abstract}        
    Previous studies have used prescriptive process monitoring to find actionable policies in business processes and conducted case studies in similar domains, such as the loan application process and the traffic fine process. However, care processes tend to be more dynamic and complex. For example, at any stage of a care process, a multitude of actions is possible. In this paper, we follow the reinforcement approach and train a Markov decision process using event data from a care process. The goal was to find optimal policies for staff members when clients are displaying any type of aggressive behavior. We used the reinforcement learning algorithms Q-learning and SARSA to find optimal policies. Results showed that the policies derived from these algorithms are similar to the most frequent actions currently used but provide the staff members with a few more options in certain situations. 
\end{abstract}

\section{Introduction}

\emph{Prescriptive} process monitoring focuses on analyzing process execution data to not only predict the future behavior of a process but also provide actionable recommendations or interventions to optimize the process~\cite{fahrenkrog2021,DBLP:conf/bpm/MetzgerKP20,DBLP:conf/caise/BozorgiDRPST23}. It goes beyond \emph{descriptive} or \emph{predictive process monitoring} by actively suggesting specific actions or decisions for improving process performance, compliance, or efficiency. Considering the decision points in business processes, the ability to offer specific guidance to users regarding optimal actions is crucial, as it can lead to improved decision-making and efficiency. 

One prominent approach is to use reinforcement learning, which learns online by interacting with an environment to adapt and improve its recommendations over time. The environments can be learned and built using the historical execution traces and the feedback they received.
%
While reinforcement learning methods have been applied in business processes, 
healthcare processes exhibit distinct characteristics and present new challenges for these techniques~\cite {DBLP:journals/jbi/Munoz-GamaMFJSH22}, such as dynamic workflows, diverse stakeholders, and patient safety considerations. In particular, patients may exhibit very diverse statuses, and a wide range of actions is possible at any stage. Moreover, each patient may react differently to these actions. These challenges may cause RL methods not to converge or not be able to improve the current policies. In such dynamic settings, it is worth investigating the validity and effectiveness of the RL approaches. 

In this paper, we focus on the healthcare domain, and in particular, the process of actions and responses in the aggression incidents by clients with intellectual impairments in community care facilities \cite{hensel_lunsky_dewa_2013}. Being in aggressive situations can have a severe impact on staff members since there is a mediation effect between experiencing aggressive behavior from clients and burnout through fear of assault \cite{mills_rose_2011}. This means that experiencing aggressive behavior leads to fear of assault, which in turn leads to burnout. It also has a negative impact on the clients themselves because aggressive behavior can lead to more aggressive behavior \cite{bushman_bonacci_pedersen_vasquez_miller_2005}. Therefore, learning the optimal way to act during aggression incidents helps de-escalate the incidents and reduce negative impact. 

Previous studies have analyzed the aggression incidents of such clients within Dutch residential care facilities using a process mining approach~\cite{agg_behav} or proposing to mine potential causal patterns~\cite{DBLP:conf/bpm/KoornLLR20,KOORN2022102035,DBLP:conf/icpm/KoornLLMVR22}. 
%
This meant that insights into the use of actions and their effects could be made visible to show which actions had a negative and which actions had a positive outcome in each situation. However, this approach can only provide recommendations for a single incident and does not take consecutive incidents and their consequences into account. 

In this paper, we investigate the use of prescriptive process monitoring, inspired by \cite{learntoact}, particularly reinforcement learning techniques, for this healthcare process, in which the optimal policies of the best possible action in a given situation (or state) can be determined. 
%
%
%
First, we train a \emph{Markov Decision Process}~(MDP) from the aggression incident log~\cite{KOORN2022102035}. Second, we apply reinforcement learning techniques, aiming to find optimal policies for staff members to minimize aggressive incidents by clients with intellectual impairments. We use the model-free, value-based control algorithms: Q-learning and SARSA. The reason for choosing these methods, rather than the Monte Carlo methods used in \cite{learntoact}, stems from their practical advantage of achieving earlier convergence on stochastic processes \cite{sutton_richard}. 

The structure of the paper is as follows. Section~\ref{sec:related} discusses the related work. Then we explain the methods in Section~\ref{sec:method}, including the description of the data set and the design of the environment. Section~\ref{sec:results} presents the results, and Section~\ref{sec:discussion} discusses the results. Section~\ref{sec:conclusion} concludes the paper.

\section{Related work}
\label{sec:related}
Research in prescriptive process monitoring has been done in the recent couple of years, mainly with a focus on business processes. Fahrenkrog-Petersen et al.~\cite{fahrenkrog2021} used it to make a framework that parameterized a cost model to assess the cost–benefit trade-off of generating alarms. Bozorgi et al.~\cite{bozorgi} researched it in the context of reducing the cycle time of general supply chain processes. Both use supervised learning methods instead of reinforcement learning methods and predict a threshold value that, when exceeded, recommends an action. The algorithms themselves do not make a recommendation; only predictions are made, and based on the predictions, a user-defined action is recommended.

Weinzierl et al.~\cite{weinzierl2020predictive} also made this remark and proposed an alternative approach to prescriptive process monitoring in which there is a learning and a recommendation phase, in which the recommendation gives the next best action to take. Branchi et al.~\cite{learntoact} used prescriptive process monitoring with Monte Carlo methods to determine the best actions to lend out loans and ensure most traffic fines are paid. The Monte Carlo methods are valid algorithms, although TD methods such as Q-learning and SARSA tend to converge earlier on stochastic processes in practice \cite{sutton_richard}. In this paper, we use Q-learning and SARSA to find optimal policies. 


\section{Methodology}
\label{sec:method}
This section describes the methods used in the research. First, we describe the data set. We then explain the preprocessing steps and the way the environment is built.  Finally, we discuss the evaluation measures used. 

\subsection{Data set}
The data set is from a Dutch residential care organization with several facilities. The event data contains 21,384 reported aggression incidents from 1,115 clients with intellectual impairments. The data has been anonymized for privacy reasons. The reported incidents were reported by staff members between the 1st of January 2015 and the 31st of December 2017.  The event data includes attributes such as the date of the incident, pseudonym client ID, the type of aggression, the countermeasure that the staff took, and the type of persons involved (such as family, staff members, and other clients). A simplified example of the event data is listed in Table~\ref{tableDesData}.

\begin{table}[b]
\centering
\caption{A snippet of the incident data where the last column describes the countermeasures taken by staff members to stop the aggression.}
\begin{tabular}{ |c|c|c|c|c| } 
 \hline
   \textbf{Pseudonym client} & \textbf{Date of incident} & \textbf{Aggression type} & \textbf{Involved} & \textbf{Measures}\\
 \hline 
     ab45 & 05/01/2016 & va & family & talk to client\\
  \hline 
     ab45 & 06/01/2016 & pp & client & none\\
  \hline 
     lz12 & 06/01/2015 & sib & unknown & seclusion\\
  \hline 
     lz12 & 18/01/2015 & po & client & none\\  
 \hline 
\end{tabular}
\label{tableDesData}
\end{table}

In the event data, four types of aggression are reported, which are \emph{verbal aggression} (va), \emph{physical aggression against people} (pp), \emph{physical aggression against objects} (po), and \emph{self-injurious behavior} (sib). Eight distinct countermeasures are reported by the staff members: \textit{talk to the client, 
held with force, no measure taken, seclusion, send to another room, distract client, terminate contact, and starting preventive measures}. 

\subsection{Data cleaning and preprocessing}
To use reinforcement learning with this dataset, we preprocess the data. We follow the same steps as in~\cite{KOORN2022102035}.
%
First, we add the type of next aggression incident as an attribute of the current event, in order to create tuples of three which contain the type of current aggression, the countermeasures taken by a staff member, and the type of next aggression. The aim is to use the aggression types as the \emph{states} a client is in and use the countermeasures as \emph{actions}. 
Such a triplet describes a transition from one state to the next state after taking an action. 

\begin{figure}
    \centering
    \begin{minipage}{1\textwidth}
        \includegraphics[width=1\textwidth]{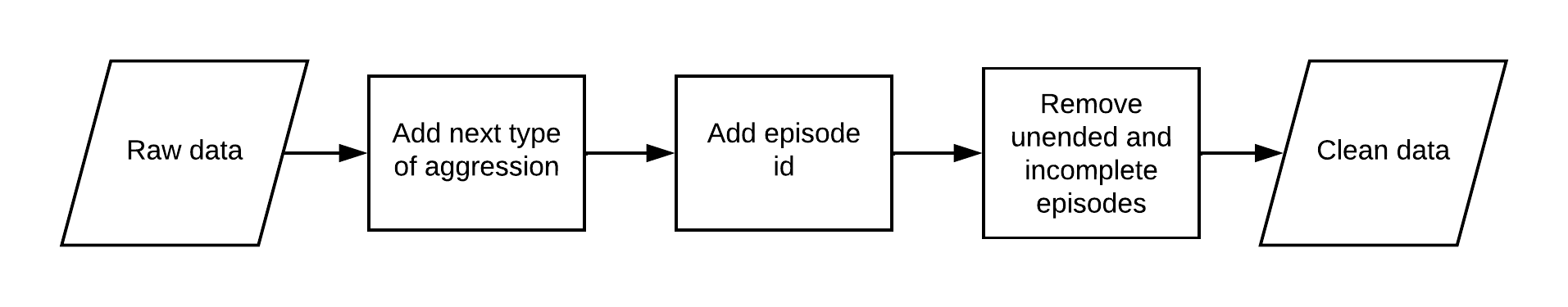}
        \caption{Preprocessing pipeline used to get enriched and clean data}
        \label{fig:pre_enrich}
    \end{minipage}\hfill
\end{figure}

In the second step, we group incidents into \emph{episodes}. According to a behavioral expert at the care organization~\cite{KOORN2022102035}, an \emph{episode} is a sequence of incidents by the same client that occurred after each other, where the time between incidents is less than or equal to nine days. 
Following this domain knowledge, we segment the sequences of incidents into episodes. When two consecutive incidents $e_i$ and $e_{i+1}$ of a client are more than nine days apart, we insert a \emph{Tau} after $e_i$ as the final state of an episode. The incident $e_{i+1}$ is the start state of the next episode. An overview of the approach is shown in Figure~\ref{fig:pre_enrich}.

We assign each episode a unique ID. The episodes that do not end in a \emph{Tau} state are considered incomplete and, therefore, filtered.  We obtained a total of 8,800 episodes after this filter, consisting of 19,848 incidents. 
%
In addition, the episodes where the incidents miss the values on the measures column are removed; these are incidents in which the staff member did not report the measures they had taken. Applying this filter reduced the number of episodes to 8,013, consisting of 15,464 incidents. 
Finally, we decided to remove the most infrequent action, `preventive measures started' due to its ambiguity and to reduce the search space. 
Any episode that contains this action was removed, resulting in 14,676 incidents and 7,812 episodes for training the final MDP. 
In Table \ref{tableFilterData}, a simplified example of the preprocessed log is listed. 

\begin{table}[ht]
\centering
\caption{A simplified example of the preprocessed event data}
\label{tableFilterData}
\resizebox{\textwidth}{!}{
\begin{tabular}{ |c|c|c|c|c| } 
 \hline
   \textbf{Pseudonym client} & \textbf{Aggression type} & \textbf{Measures} & \textbf{Next aggression type} & \textbf{Episode Id}\\
 \hline 
     ab45 & va & talk to client & pp & 1\\
  \hline 
     ab45 & pp & none & Tau & 1\\
  \hline 
     lz12 & sib & secluded & Tau & 2\\
  \hline 
     lz12 & po & none & Tau & 3\\  
 \hline 
\end{tabular}
}
\end{table}

\subsection{Building the environment}
Now that the data is cleaned and preprocessed, we use it to build a finite MDP. For this, we need the five-tuple consisting of the states, actions, transition probabilities, rewards, and discount factor \cite{sutton_richard}. The discount factor is a hyperparameter that can be tuned; therefore, we later perform hyperparameter tuning to determine the discount factor for the agent.

We describe the MDP using the standard formalization in \cite{sutton_richard} as follows:
\begin{itemize}
    \item $\mathcal{S}$ = \{va, po, sib, pp, Tau\}, i.e., the set of states;
    \item $\mathcal{A}$ = $\{$talk to the client, no measure taken, seclusion, holding with force, send to another room, distract client, terminate contact$\}$, i.e., the set of actions;
    \item $\mathcal{P}$, which is the probability of going from one state to the next based on the action. This is determined using the following function \begin{equation}
        P(s,a,s')= \frac{\textrm{Number of times } a \textrm{ leads to } s'}{\textrm{Number of times } a \textrm{ is chosen in state } s}
    \end{equation}
    \item $\mathcal{R} : \mathcal{A} \times \mathcal{S} \rightarrow \mathbb{Z}$, which is the reward function. We defined the reward function based on the literature in assessing the severity of the action and the state~\cite{soas-r}. The reward (penalty) for each individual action or state is listed in Table \ref{tab:reward_table}. 
    \begin{table}[tb]
        \centering
         \caption{The rewards (penalty) assigned for each action or state, based on the severity of the action and the state~\cite{soas-r}. When an agent takes an action and ends in the follow-up state, the combination of the action and state is used to compute the reward.}
        \label{tab:reward_table}
        \begin{tabular}{|l|c|}
        \hline
            \textbf{Action or state} & \textbf{Reward}\\
        \hline
            Tau & 1\\
        \hline
            Verbal Aggression (va) & 0\\
        \hline
            Physical Aggression against objects (po) & -1\\
        \hline
            Self-injurious behavior (sib) & -3\\
        \hline 
            Physical Aggression against people (pp) & -4\\
        \hline
            Client distracted, Contact terminated, Send to other room & -1\\
        \hline
            Hold with force, Seclusion & -2\\
        \hline
            Other actions & 0\\
        \hline
        \end{tabular}
    \end{table}
\end{itemize}

\begin{figure}[ht]
    \centering
        \includegraphics[width=0.7\textwidth]{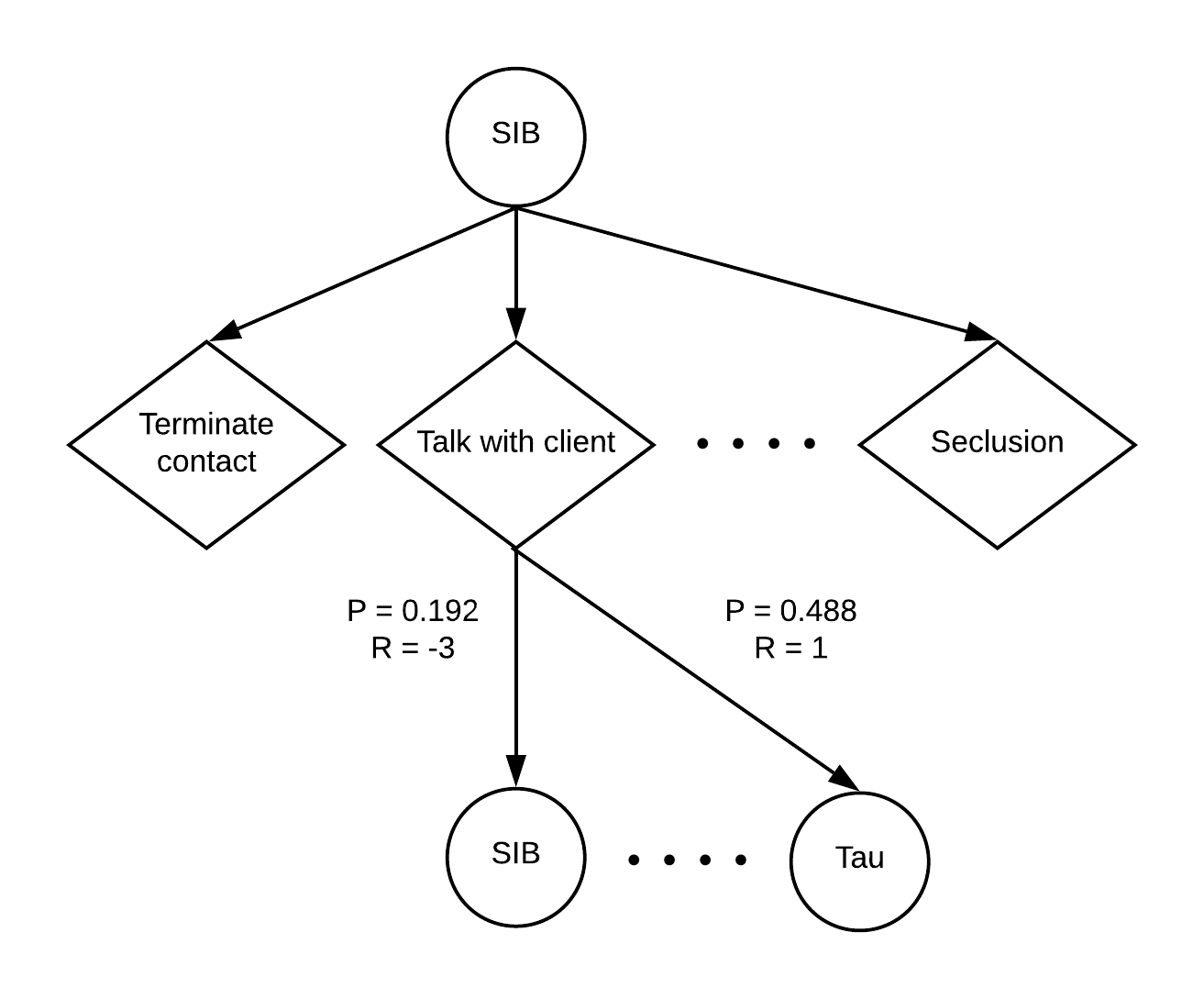}
        \caption{A subgraph of the MDP, depicting the current state of self-injurious behavior (SIB), a sample of actions that can be chosen, and a sample of transitions. $P$ is the probability of going to that state, and $R$ is the reward associated with that action and next state.}
        \label{fig:mdp_sib}
\end{figure}

Another design choice has been made regarding the calculation of the transition probabilities. In the data set, multiple actions could be filled in at each incident. For this paper, a decision was made to consider only the most frequent action as the transition from one state to the next, in order to limit the number of possible actions and avoid having too many infrequent actions. Also, the reward function was designed based on the severity of the action and the state as indicated in the existing literature in aggression~\cite{soas-r}. The simple reward function was designed on purpose such that the results can be more easily communicated to the experts.  
A subgraph of the environment can be seen in Figure \ref{fig:mdp_sib}.

\subsection{Training the agents}\label{sec:hypertune}
We used the following parameters in the tuning: the learning rate, $\alpha \in [0,1]$, the discount factor, $\gamma \in [0,1]$, and the amount of exploration, $\epsilon \in [0,1]$, which have an impact on the training of the agents and therefore the results. The best parameters are obtained experimentally by hyperparameter tuning using the best average reward of 100 runs as the goal, each consisting of 2000 episodes. The search spaces are $\alpha \in [0.1, 0.2, 0.3, 0.4, 0.5]$, $\gamma\in[0.2, 0.4, 0.6, 0.8, 1.0]$ and $\epsilon\in[0.1, 0.2, 0.3, 0.4, 0.5]$. First $\gamma$ was obtained while keeping $\alpha$ and $\epsilon$ on 0.1. After this $\epsilon$ was obtained using the optimal $\gamma$ value and $\alpha=0.1$ and finally $\alpha$ was obtained using the optimal $\gamma$ and $\epsilon$ value. Each parameter has been used for ten different runs to get a fair average. The hyperparameter values for both Q-learning and SARSA are $\alpha=0.2$, $\gamma=0.2$ and $\epsilon=0.1$.

\subsection{Evaluation of policies}
We evaluate the agents both quantitatively, by comparing the average rewards, and qualitatively, by discussing the policies. For the quantitative evaluation, we compute the average reward for the best-trained agent using Q-learning and the best-trained agent using SARSA. These are then compared with the average reward of taking random actions and the average reward with the current policy. The current policy has been derived as the most frequent action taken in a state. The current policy is ``talking to the client'' when they display verbal aggression (va), physical aggression against people (pp), and physical aggression against objects (po). For the state of self-injurious behavior (sib) ``no action'' is the most frequently used action. For the qualitative evaluation, we discuss the results by looking at the most frequent variants for each agent and comparing these variants with the ones of the current policy. 

\section{Results}
\label{sec:results}
In this section, we first present the results regarding the rewards. Next, we discuss the results qualitatively, presenting the optimal policy and the variants. 
We used two baselines to compare the results: (1) using random actions and (2) taking the most frequent action at each state as the policy for the agent. The data set is shared under the NDA and thus unavailable. The code and the MDPs used in this paper are online available\footnote{  \url{https://git.science.uu.nl/6601723/ppm-aggressive-incidents}}, which can be used to reproduce the results. 

\subsection{Quantitative results}
In this section, the average reward per policy is described and evaluated. It is listed in Table~\ref{tableAvgRew}.

\begin{table}[ht]
\centering
\caption{Average reward per policy based on 10,000 runs, each with 100 episodes.}
\label{tableAvgRew}
\begin{tabular}{ |c|c| } 
 \hline
   \textbf{Policy} & \textbf{Average reward} \\
 \hline 
     Random & -3.783 \\
  \hline 
     Most frequent action & -1.105 \\
  \hline 
     Q-learning & -1.127 \\
  \hline 
     SARSA & -1.168 \\  
 \hline 
\end{tabular}
\end{table}

We run each policy for 10,000 runs, each consisting of 100 episodes, resulting in 1,000,000 episodes total. To test if the differences between the policies are significant, we performed a one-way ANOVA with the data from SARSA, Q-Learning, and the current policy. The one-way ANOVA was done using the Scipy library from Python 3, specifically the stats.f\_oneway function. The p-value was $7.719e-26$, which is smaller than 0.05, therefore, we can reject the null hypothesis that the groups have the same mean, meaning there is a significant difference between them, but we do not know between which. Therefore, we use a least significant difference test as a posthoc test using the pairwise\_tukeyhsd from the Python 3 library statsmodels. This made three comparisons, it tested Q-learning against the current policy, tested SARSA against the current policy, and tested Q-learning against SARSA. It rejected all three null hypotheses, meaning the average reward per algorithm significantly differs from one another. 

\subsection{Qualitative results}
This section describes the qualitative results where we show the derived policies and the most common variants of episodes per agent. The derived policies can be found in Table \ref{tablePolicy}, where the action taken at each state for each policy can be found. 

\begin{table}[ht]
\centering
\caption{Derived policies for Q-learning and SARSA together with the most frequent actions taken on the 10,000 runs, each with 100 episodes.}
\label{tablePolicy}
\begin{tabular}{ |c|c|c|c|c| } 
 \hline
   \textbf{Policy} & \textbf{VA} & \textbf{SIB} & \textbf{PP} & \textbf{PO}\\ 
  \hline 
     Most frequent action & talk with client & no measure & talk with client & talk with client \\
  \hline 
     Q-learning &  talk with client & no measure & talk with client & no measure\\
  \hline 
     SARSA & no measure & no measure & talk with client & talk with client \\  
 \hline 
\end{tabular}
\end{table}

The five most common variants with their frequencies for each of the agents can be found in Tables \ref{tablePathsMostFreq}, \ref{tablePathsQ}, and \ref{tablePathsSARSA}. In the tables, each variant is a distinct episode of tuples, where the first element of the tuple is the \emph{current state}, the second element is the \emph{action} taken, and the last element is the \emph{next state} after the action. If the state is \emph{Tau}, the episode is ended; otherwise, another action is taken.


\begin{table}
\centering
\caption{Five most common }
\label{tablePathsMostFreq}
\resizebox{0.7\textwidth}{!}{
\begin{tabular}{ |c|c| } 
 \hline
   \textbf{Path} & \textbf{Frequency} \\
 \hline 
    (va, Talk with client, Tau) & 14454 \\
    \hline
    (sib, No measure, 'Tau') & 13987 \\
    \hline
    (po, Talk with client, Tau) & 13100 \\
    \hline
    (pp, Talk with client, Tau) & 12769 \\
    \hline
    (pp, Talk with client, pp)(pp, Talk with client, Tau) & 4454 \\
 \hline 
\end{tabular}
}
\end{table}

\begin{table}
\centering
\caption{Five most common variants when using the policy derived by Q-learning.}
\label{tablePathsQ}
\resizebox{0.7\textwidth}{!}{
\begin{tabular}{ |c|c| } 
 \hline
   \textbf{Path} & \textbf{Frequency} \\
 \hline 
  (va, Talk with client, Tau) & 14417 \\
  \hline
  (sib, No measure, Tau) & 14294 \\
  \hline 
  (po, No measure, Tau) & 12974 \\
  \hline 
  (pp, Talk with client, Tau) & 12866 \\
  \hline 
  (pp, Talk with client, pp)(pp, Talk with client, Tau) & 4526 \\
 \hline 
\end{tabular}
}
\end{table}

\begin{table}
\centering
\caption{Five most common variants when using the policy derived by SARSA.}
\label{tablePathsSARSA}
\resizebox{0.65\textwidth}{!}{
\begin{tabular}{ |c|c| } 
 \hline
   \textbf{Path} & \textbf{Frequency} \\
 \hline 
    (va, No measure, Tau) & 14926 \\
    \hline 
    (sib, No measure, Tau) & 14025 \\
    \hline 
    (pp, Talk with client, Tau) & 13079 \\
    \hline 
    (po, Talk with client, Tau) & 12971 \\
    \hline 
    (sib, No measure, sib)(sib, No measure, Tau) & 4313 \\
 \hline 
\end{tabular}
}
\end{table}

In the tables, it can be seen that the four most frequent variants of episodes end in one action for all policies. For each state doing that action leads immediately to Tau regardless of the policy. Also, most of the episodes ended when only one action had been taken. 
When we take a closer look at the current policy, the Q-learning policy, and the SARSA policy, we see that most variants are the same with only two differences: (1)~in the verbal aggression state (va), ``no measure'' action is suggested by the Q-learning; (2)~in the self-injury-behavior state (sib), ``no measure'' action is suggested by SARSA.

\subsection{Additional analysis}
Due to the results from taking all episodes, we decided to do an additional analysis on a subset of the data. We kept the episodes that had a length of more than or equal to three incidents and performed the same experiment as we did on the whole dataset. This subset contained 6687 incidents over 1360 episodes. Taking only the episodes longer than or equal to three incidents, we focus on the clients who display more severe behavior, which are the ones we want to help reduce in the first place. We again used Q-learning and SARSA as described above and compared them to taking random actions and the current policy, which in this case was ``talk with client'' in every state. The hyperparameter tuning was done the same as described in Section~\ref{sec:hypertune}, resulting in the best performing Q-learning agent and best performing SARSA agent both with $\alpha=0.1$, $\gamma=0.2$ and $\epsilon=0.1$.
In the remaining parts of the additional analysis, we present the quantitative and qualitative results.

\subsubsection{Additional analysis quantitative results}
In this section, the average reward per policy is shown and can be found in Table~\ref{tableAvgRewAdd}.

\begin{table}[ht]
\centering
\caption{Average reward per policy based on 10000 runs along 100 episodes}
\label{tableAvgRewAdd}
\begin{tabular}{ |c|c| } 
 \hline
   \textbf{Policy} & \textbf{Average reward} \\
 \hline 
     Random & -11.925 \\
  \hline 
     Most frequent action & -7.342 \\
  \hline 
     Q-learning & -7.266 \\
  \hline 
     SARSA & -7.275 \\  
 \hline 
\end{tabular}
\end{table}

The same statistical tests as on the whole dataset were done. The p-value of the one-way ANOVA was $2.917e-12$, which is smaller than 0.05, therefore, we can reject the null hypothesis. We use a least significant difference test as a posthoc test. This made three comparisons. It rejected two out of three null hypotheses. With a p-adj value of 0.6833, it did not reject the hypothesis that the rewards from Q-Learning and SARSA had the same mean. When taking a look at Table~\ref{tablePolicyAdd}, we can see that they have learned exactly the same policy, so this result was expected. 

\subsubsection{Additional analysis qualitative results}
We list the derived policies with the current policy and compare the most common variants taken between a random baseline, the most frequent actions taken, and the two policies derived by the agents.

The derived policies can be found in Table~\ref{tablePolicyAdd}, where the action taken at each state for each policy can be found. The Q-learning and SARSA agent learned that ``talking to a client'' was the best option when the state is verbal aggression (va), physical aggression against people (pp), or physical aggression against objects (po), and ``no measure'' when the state is self-injury behavior (sib).

\begin{table}[ht]
\centering
\caption{Derived policies for Q-learning and SARSA together with the most frequent actions taken on a subset of the data set only containing episodes $\geq 3$}
\label{tablePolicyAdd}
\resizebox{0.9\textwidth}{!}{
\begin{tabular}{ |c|c|c|c|c| } 
 \hline
   \textbf{Policy} & \textbf{VA} & \textbf{SIB} & \textbf{PP} & \textbf{PO}\\ 
  \hline 
     Most frequent action & talk with client & talk with client & talk with client & talk with client \\
  \hline 
     Q-learning &  talk with client & no measure & talk with client & talk with client\\
  \hline 
     SARSA & talk with client & no measure & talk with client & talk with client \\  
 \hline 
\end{tabular}
}
\end{table}

The five most common variants with their frequencies for the current policy and the RL agents can be found in Tables 
\ref{tablePathsMostFreqAdd} and \ref{tablePathsQAdd}. Both the Q-learning agent and SARSA agent learned the same policy. One noticeable difference between the frequent episodes of the policies of the RL agents and the most frequent policy is that the second most frequent episode of self-injurious behavior is added in Tables \ref{tablePathsMostFreqAdd} and \ref{tablePathsQAdd}.


\begin{table}
\centering
\caption{Five most common variants of episodes when using the most frequent actions for 1,000,000 episodes, plus the most common variant when considering SIB.}
\label{tablePathsMostFreqAdd}
\resizebox{0.7\textwidth}{!}{
\begin{tabular}{ |c|c| } 
 \hline
   \textbf{Path} & \textbf{Frequency} \\
 \hline 
    (va, Talk with client, Tau) & 5052 \\
    \hline
    (sib, Talk with client, Tau) & 4772 \\
    \hline
    (pp, Talk with client, Tau) & 4520 \\
    \hline
    (po, Talk with client, Tau) & 4488 \\
    \hline
    (pp, Talk with client, pp)(pp, Talk with client, Tau) & 2617 \\
    \hline
    (sib, Talk with client, sib)(sib, Talk with client, Tau) & 1776 \\
 \hline 
\end{tabular}
}
\end{table}

\begin{table}
\centering
\caption{Five most common variants by Q-learning for 1,000,000 episodes, plus the most common variant when considering SIB.}
\label{tablePathsQAdd}
\resizebox{0.7\textwidth}{!}{
\begin{tabular}{ |c|c| } 
 \hline
   \textbf{Path} & \textbf{Frequency} \\
 \hline 
    (va, Talk with client, Tau) & 5207 \\
    \hline
    (sib, No measure, Tau) & 4901 \\
    \hline
    (po, Talk with client, Tau) & 4661 \\
    \hline
    (pp, Talk with client, Tau) & 4620 \\
    \hline
    (pp, Talk with client, pp)(pp, Talk with client, Tau) & 2749 \\
    \hline
    (sib, No measure, sib)(sib, No measure, Tau) & 2720 \\
 \hline 
\end{tabular}
}
\end{table}

In the tables, it can be seen that the four most frequent variants end in one action for all policies. For each state, performing that action leads immediately to Tau regardless of the policy. This time only a fifth of all episodes lead immediately to Tau, for all policies except the random one. The random agent only has episodes ending with only one incident in its top ten, but this can be explained by their frequency, which is relatively much lower than the other policies (976 vs. 5052). When we take a closer look at the current policy and the RL agents' learned policy, the variants differ significantly in the frequencies of the self-injurious behavior (sib) state. The frequencies of the single-incident episodes for this state are similar between the RL agents and the current policy (4901 vs. 4772). 
When the episodes consist of two incidents that concern the self-injurious behavior (sib) state, the frequency of such episodes is much higher in the RL agents learned policy than the current policy (2720 vs. 1776), meaning that ``no measure''  results faster to Tau than ``Talk with client'' in this case.

\section{Discussion}
\label{sec:discussion}
The results indicate that the current policy and the RL-derived policies reach similar conclusions. The current policy performs slightly better than the RL agents when considering all episodes, but the RL agents provide staff members with additional options without having a significant negative impact on rewards. When considering the selected subset of the episodes, the RL agents slightly outperform the current policy, offering an alternative choice. 

In both cases, the staff member can choose to talk to the client or take no action. Although the RL algorithms exhibit slight variations in performance compared to the current policy, the policies derived do not significantly differ. This alignment is reasonable considering the reward function used, which penalizes all actions except ``no measure'' and ``talk with client''. These options align with the least disruptive impact on both the client and staff member, as indicated by previous studies.

However, it is important to note that the models may oversimplify the real situation, and further factors such as location, time, and individuals involved have not been included. Collecting relevant data and consulting behavioral experts could enhance future research in this field. For example, it is possible to learn the time distribution until the next incident and use this in the reward function. 

Additionally, practical relevance should be acknowledged, as staff members face challenges in assessing situations and may need to use force in certain cases. Future research may aim to provide insights tailored to specific clients or client groups. Combining reinforcement learning and process mining in prescriptive process monitoring shows promise but requires careful consideration of data availability and exploration limitations.

\section{Conclusion}
\label{sec:conclusion}
This paper presents the application of reinforcement learning (RL) to optimize response policies in healthcare processes, specifically addressing aggressive incidents in care settings. The research aims to investigate the validity of RL in healthcare and the ability to find optimal response policies for staff members towards such incidents.
The results have shown that RL algorithms can find such an optimal policy, which consists of taking no measures or talking with the client depending on the state. 
The policies are very similar to the current policy, i.e., the most frequent action taken by staff members. 

%

Despite the simple MDP, the results do show that prescriptive process monitoring can be used in the healthcare domain. Interestingly, it may be more beneficial to use the techniques in more complex situations, rather than the simple situation. However, further research is necessary to validate this finding. 

For future work, one may refine the environment by extending the MDP with more refined states and actions. Future research should be multidisciplinary, where such an environment can be more elaborately built based on experts in the field of aggressive behavior and staff members who work daily with clients. Results can then also be validated by the experts or staff to help them make better decisions and therefore their input is crucial.

{\small
\subsubsection*{Acknowledgement} This research was supported by the NWO TACTICS project (628.011.004) and Lunet in the Netherlands. We would like to thank the experts from the Lunet for their assistance. We also thank Dr. Shihan Wang and Dr. Ronald Poppe for the invaluable discussions.}




\bibliographystyle{splncs}
\bibliography{references.bib}
\end{document}